\documentclass[11pt]{article}

\usepackage[preprint]{acl}

\usepackage{times}
\usepackage{latexsym}
\usepackage{amsmath}
\usepackage[most]{tcolorbox}
\usepackage[table]{xcolor}
\usepackage{graphicx}
\usepackage{booktabs}
\usepackage{longtable}
\usepackage{multirow}
\usepackage[T1]{fontenc}
\usepackage[utf8]{inputenc}

\usepackage{microtype}

\usepackage{inconsolata}

\usepackage{enumitem} % Add to the preamble
\usepackage{graphicx}
\usepackage{adjustbox}
\title{Can Jev Improve Demonstration Retrieval for In-Context Learning?}

\author{
\textbf{Rui Song\textsuperscript{1}},
\textbf{Yingji Li\textsuperscript{1}},
\textbf{Hao Xu\textsuperscript{1,*}}
\\
\\
\textsuperscript{1}College of Computer Science and Technology, Jilin University, Changchun, China
\\
\small{
\textbf{*Corresponding author:} Hao Xu (xuhao@jlu.edu.cn)
}
}

\begin{document}
\maketitle
\begin{abstract}
Demonstration selection plays a crucial role in the effectiveness of in-context learning (ICL) for large language models (LLMs). Extensive research has explored various retrieval strategies to identify informative demonstrations and improve downstream performance. Recently, the emergence of Jev, a decision-oriented language model capable of producing probabilistic judgments, offers a new perspective on evaluating demonstration utility. This motivates us to investigate a fundamental question: Can Jev improve demonstration retrieval for in-context learning? In this paper, we demonstrate that Jev can effectively enhance demonstration retrieval by directly assessing the utility of candidate demonstrations for a given query. However, applying Jev to large-scale retrieval introduces substantial computational overhead, as each candidate requires an additional decision-making inference. To address this limitation, we propose a simple yet effective knowledge distillation approach that transfers Jev's utility estimation capability to a lightweight multilayer perceptron (MLP). By learning to approximate Jev's probabilistic judgments, the distilled model enables efficient demonstration ranking while maintaining competitive retrieval performance. Our study highlights the potential of decision-oriented language models for demonstration selection and provides an efficient framework for integrating their decision-making capabilities into ICL.
\end{abstract}

\section{Introduction}
In-context learning (ICL) has emerged as a remarkable capability of large language models (LLMs), enabling them to adapt to diverse downstream tasks without requiring parameter updates~\cite{song2024shortcut}. Demonstrations serve as the foundation of ICL, allowing LLMs to infer task-relevant patterns and input–output relationships from a few labeled examples~\cite{brown2020language, song2026retrieval}. Consequently, extensive research has explored various demonstration selection strategies to identify informative examples and improve ICL performance.

Existing demonstration selection methods can be broadly categorized into heuristic-based and training-based approaches. Heuristic-based methods typically quantify the relevance between candidate demonstrations and a given query using lexical matching, such as BM25, semantic similarity in embedding spaces or perplexity, and select the highest/lowest-scoring candidates as demonstrations \cite{liu2022makes, robertson2009probabilistic, xu2026toward}. To further improve selection quality, some approaches adopt a multi-stage retrieval paradigm, where an initial retrieval stage narrows the candidate pool based on static relevance scores, followed by further refinement that incorporates demonstration diversity, redundancy reduction, or feedback from LLMs \cite{hongjin2022selective, peng2024revisiting, song2026fake}. In contrast, training-based approaches learn task-aware retrievers by leveraging supervision signals derived from LLM predictions, likelihoods, or performance feedback to identify more informative demonstrations~\cite{zhang2025learning,wang2026easier}. 

Despite their effectiveness, heuristic-based methods often struggle to capture the actual utility of demonstrations, as predefined relevance metrics do not necessarily reflect their contribution to downstream predictions. Although training-based methods can better estimate demonstration utility through LLM feedback, they typically incur substantial computational costs and require retraining for different tasks or target models, limiting their scalability and cross-task generalizability. 

Motivated by the recent emergence of Jev~\cite{typesafe2026jev}, a decision model capable of making probabilistic judgments, we investigate whether its decision making capabilities can improve demonstration retrieval for ICL. Specifically, we first employ heuristic retrieval to narrow the candidate pool and then leverage Jev to rerank demonstrations according to their estimated utility for a given query. Our experiments show that this simple strategy yields substantial performance improvements. However, repeatedly evaluating candidate demonstrations with Jev introduces considerable computational overhead and inference latency. To address this limitation, we propose a task agnostic distillation framework that transfers Jev's utility estimation capability to a lightweight multilayer perceptron (MLP). By learning to approximate Jev's judgments of demonstration utility, the MLP efficiently refines retrieved candidates without requiring repeated Jev inference. Experimental results demonstrate that our approach achieves a favorable balance between effectiveness and efficiency, while exhibiting strong generalizability across diverse tasks with a single distillation process.
Our main contributions are summarized as follows:

\begin{itemize}[leftmargin=*, nosep]
    \item We investigate the potential of Jev for ICL demonstration retrieval and demonstrate that its probabilistic judgments can effectively identify informative demonstrations.

    \item We propose a simple and task-agnostic distillation framework that transfers Jev's demonstration utility estimation capability to a lightweight MLP, reducing computational overhead while improving generalizability across tasks.

    \item Experiments across diverse tasks and LLMs demonstrate the effectiveness of our approach, achieving strong retrieval performance, improved efficiency, and robust cross-task generalization.
\end{itemize}

\section{Related Work}

\paragraph{Demonstration Retrieval in ICL. }
Learning from demonstrations has emerged as a fundamental capability of LLMs, prompting extensive research on improving their performance through more effective demonstration retrieval strategies. Early studies primarily relied on heuristic-based retrieval methods, such as lexical matching and semantic similarity, to identify demonstrations relevant to a given query \cite{robertson2009probabilistic,liu2022makes}. Furthermore, several studies extend efficient embedding-based retrieval by incorporating additional signals, such as perplexity, prediction confidence, and demonstration diversity, to refine candidate selection \cite{iter2023context,peng2024revisiting,yang2023representative,xu2026toward}. But these methods often rely on predefined heuristics that may not reflect actual demonstration utility, limiting their effectiveness and adaptability across LLMs.

Alternatively, training-based methods partially address these limitations by learning demonstration retrievers from LLM feedback. EPR \cite{rubin2022learning} trains a dense retriever using demonstration utility estimated from output probabilities. UDR \cite{li2023unified} further introduces multi-task list-wise ranking to learn a unified retriever across tasks. Wang et al. \cite{wang2024learning} employ reward modeling and knowledge distillation to iteratively optimize retrieval. GenICL \cite{zhang2025learning} leverages generative preference learning to directly optimize demonstration selection based on LLM feedback. More recently, DiSP \cite{wang2026easier} trains lightweight judges to predict the success of candidate demonstration sets, enabling efficient selection without exhaustive search. Despite their effectiveness, these methods typically require substantial LLM feedback collection and additional training, increasing the cost of adapting to new tasks or models. 

\paragraph{Recent Research in Jev.}
The recent emergence of Jev, a specialized decision model, has attracted increasing attention by reducing the computational cost of LLM-based decision making. Recent studies have investigated its effectiveness and reliability across diverse tasks. Several benchmarking studies have demonstrated its competitive performance in text classification, natural language inference, and other decision-oriented tasks \cite{deusser2026evaluating,li2026specialized}. Meanwhile, Jev-as-a-Judge \cite{li2026jev} explores confidence-aware decision making by selectively invoking more powerful LLMs for uncertain predictions, while other studies examine the calibration and reliability of Jev's probabilistic outputs \cite{meng2026far}. More recently, General Decision Models \cite{duan2026general} investigates the broader capabilities of decision models and explores efficient decision making through knowledge distillation. Despite these advances, whether Jev's probabilistic judgments can effectively guide demonstration retrieval, particularly through lightweight and task-agnostic reranking, remains underexplored.

\section{Preliminary Experiment}

\subsection{Problem Formulation}

\paragraph{In-Context Learning.}
Given a labeled dataset $\mathcal{D}=\{(x_i,y_i)\}_{i=1}^{N}$ and a target query $x_q$, ICL enables an LLM $f_\theta$ to predict the target label without updating its parameters. Specifically, a set of $K$ labeled demonstrations $\mathcal{C}_q=\{(x_i,y_i)\}_{i=1}^{K}\subseteq\mathcal{D}$ is incorporated into the input context, and the prediction is $\hat{y}_q=\arg\max_{y\in\mathcal{Y}}P_\theta(y\mid \mathcal{C}_q,x_q),$
where $\mathcal{Y}$ denotes the label space and $P_\theta$ represents the conditional predictive distribution of the LLM.

\paragraph{Demonstration Retrieval.}
Based on the above definitions, a retrieval method selects $K$ demonstrations according to a scoring function $s$:
\begin{equation}
\mathcal{C}_q^{s}=\operatorname{TopK}_{(x_i,y_i)\in\mathcal{D}}s(x_q,x_i).
\end{equation}
Existing methods typically define $s$ based on lexical or semantic similarity, which may not accurately reflect the utility of demonstrations for the target prediction.

\subsection{Jev-guided Reranking}

Jev is a specialized decision model that estimates the probabilities of candidate decisions given an input context. This capability enables us to evaluate whether a labeled demonstration $(x_i,y_i)$ is potentially beneficial for predicting a target query $x_q$. Specifically, we formulate demonstration utility estimation as a binary decision problem, where Jev assesses the usefulness of each candidate through the following instruction:

\begin{tcolorbox}[
    enhanced,
    breakable,
    colback=gray!3,
    colframe=gray!45,
    colbacktitle=gray!15,
    coltitle=black,
    fonttitle=\bfseries\small,
    title={Jev Prompt},
    boxrule=0.5pt,
    arc=2pt,
    left=8pt,
    right=8pt,
    top=7pt,
    bottom=7pt,
    boxed title style={
        boxrule=0pt,
        arc=2pt
    }
]
\small

\textbf{Classification Task:} \texttt{[Task description]}

\medskip
\textbf{Query to Solve:} \texttt{[Target query]}

\medskip
\textbf{Candidate Demonstration:} \texttt{[Candidate text]}\\
\textbf{Candidate Label:} \texttt{[Candidate label]}

\medskip
Would this labeled candidate be a useful and relevant
in-context demonstration for correctly classifying
the query?

\medskip
\texttt{[options]} \texttt{false} \texttt{true}
\\
\texttt{[decision]:}
\end{tcolorbox}

Given the instruction $I$, we define the Jev-based utility score as
\begin{equation}  \label{eq:jev_score}
s_{\mathrm{Jev}}(x_q,x_i,y_i)
=P_{\mathrm{Jev}}(\mathrm{Yes}\mid I,x_q,x_i,y_i).
\end{equation}
A higher score indicates that Jev considers the candidate more likely to be helpful for the prediction.

Although Jev enables efficient decision making, evaluating every candidate in the entire pool $\mathcal{D}$ remains computationally expensive. Inspired by existing demonstration retrieval methods, we therefore adopt a two-stage retrieval framework \cite{peng2024revisiting,xu2026toward}. First, we employ a pretrained Sentence-Transformer encoder $g$ to retrieve the top-$M$ semantically similar candidates:
\begin{equation}
\mathcal{R}_q=\operatorname{TopM}_{(x_i,y_i)\in\mathcal{D}}
\operatorname{sim}(g(x_q),g(x_i)),
\label{eq:semantic_retrieval}
\end{equation}
where $\operatorname{sim}(\cdot,\cdot)$ denotes cosine similarity. In the second stage, Jev evaluates the utility of each retrieved candidate and reranks them according to Eq.~\eqref{eq:jev_score}. The final $K$ demonstrations are selected as
\begin{equation}
\mathcal{C}_q^{\mathrm{Jev}}
=\operatorname{TopK}_{(x_i,y_i)\in\mathcal{R}_q}
s_{\mathrm{Jev}}(x_q,x_i,y_i),
\label{eq:jev_reranking}
\end{equation}
where $K\leq M$. This two-stage framework combines the efficiency of semantic retrieval with the decision-making capability of Jev, enabling utility-aware demonstration selection without evaluating the entire candidate pool.

\subsection{Preliminary Experimental Results} 
\begin{figure}
    \centering
    \includegraphics[width=1\linewidth]{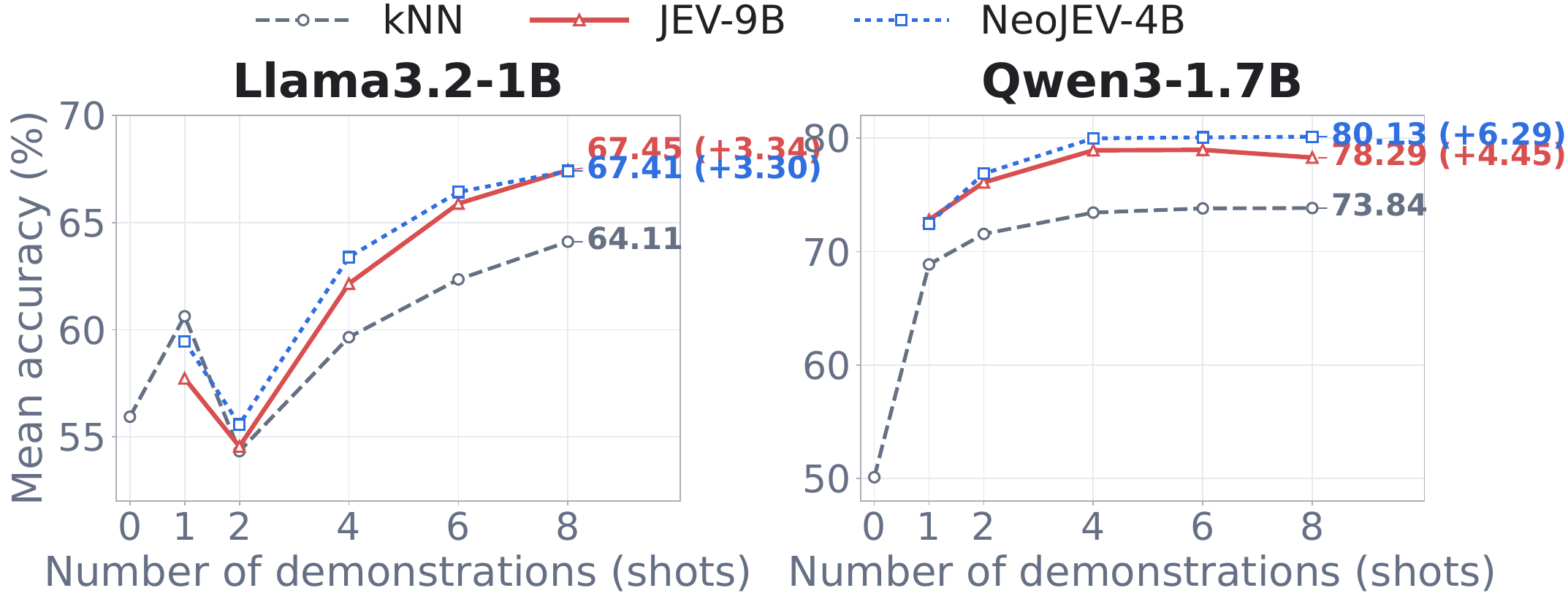}
    \caption{Average results over five datasets with the increasing of $K$. More detailed results are in App. \ref{app:more_r1}.}
    \label{fig:jev_icl_avg}
\end{figure}

To investigate the effectiveness of Jev-guided reranking, we set the candidate pool size to $M=20$ and conduct experiments using two backbone LLMs, Llama3.2-1B and Qwen3-1.7B. We employ two open-source decision models, NeoJev (4B)\footnote{https://huggingface.co/TokenRhythm/NeoHorse-Jev-4B} and JEV (9B)\footnote{https://huggingface.co/autotrust/JEV-9B}, and report the average accuracy across five datasets: TREC for question classification, SST-2 and SST-5 for binary and fine-grained sentiment classification, respectively, AG News for topic classification, and MNLI for natural language inference as used in \cite{wang2026easier}. Further details on the datasets and experiment are provided in App. \ref{app:details}. The results are presented in Fig.~\ref{fig:jev_icl_avg}.

JEV-guided retrieval generally improves ICL performance over conventional $k$NN retrieval across different numbers of demonstrations \cite{liu2022makes}. As the number of demonstrations increases, both JEV-9B and NeoJEV-4B tend to achieve larger improvements. At 8-shot, JEV-9B improves the average accuracy by 3.34 and 4.45 percentage points on Llama3.2-1B and Qwen3-1.7B, respectively, while NeoJEV-4B achieves gains of 3.30 and 6.29 percentage points. Notably, despite having fewer parameters, NeoJEV-4B achieves comparable performance on Llama3.2-1B and substantially outperforms JEV-9B on Qwen3-1.7B. This observation is consistent with the stronger discriminative capabilities reported for NeoJEV-4B in its official evaluation, suggesting that decision-making quality, rather than model size alone, is critical for identifying informative demonstrations. Although the improvements are not strictly monotonic across shots, these findings demonstrate the potential of decision-oriented models to guide effective demonstration selection beyond semantic similarity.

\subsection{Quantitative Analysis of Jev Retrieval}

\begin{figure}
    \centering
    \includegraphics[width=1\linewidth]{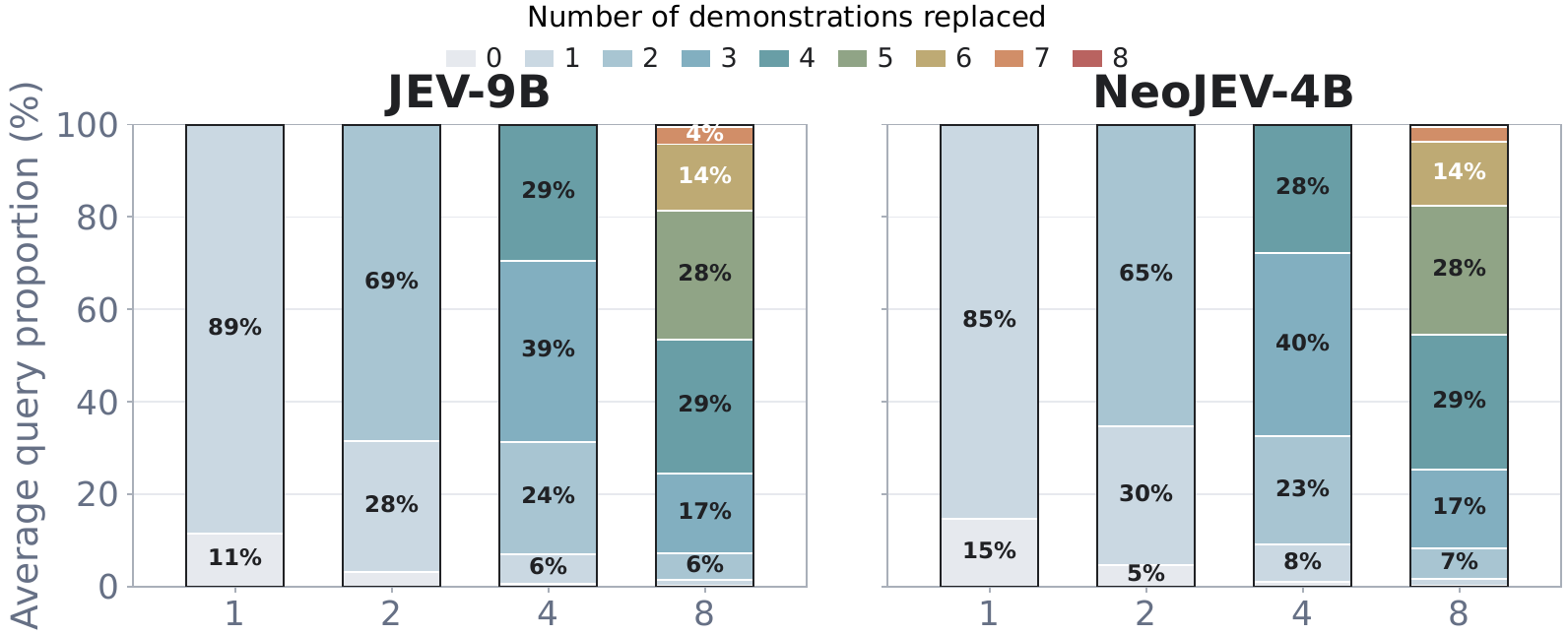}
    \caption{Distribution of demonstration replacements.}
    \label{fig:jev_ratio}
\end{figure}

\begin{figure}
    \centering
    \includegraphics[width=1\linewidth]{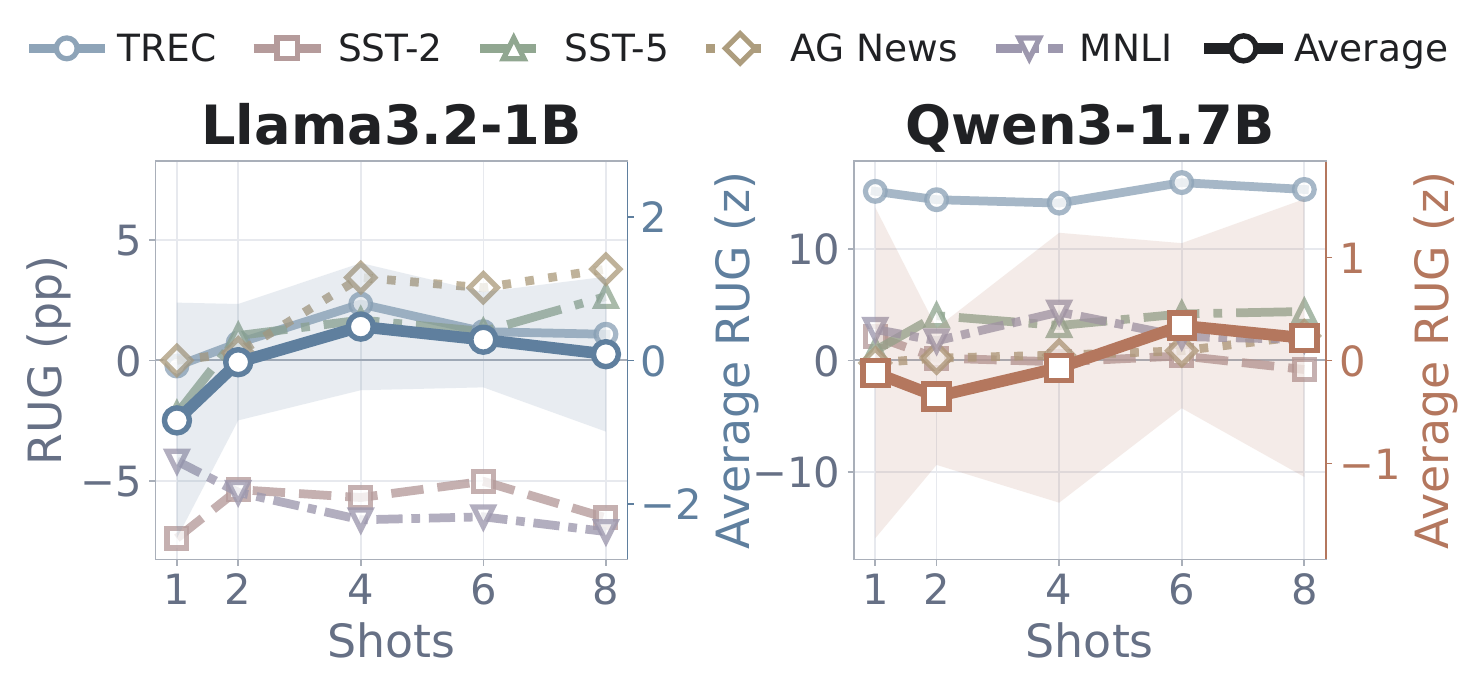}
    \caption{RUG scores across datasets and the average RUG results. }
    \label{fig:jev_qua}
\end{figure}

Although end-to-end ICL accuracy demonstrates the overall effectiveness of JEV-guided retrieval, it provides limited insight into how JEV modifies the demonstrations retrieved by kNN and whether the newly selected demonstrations are more useful than those they replace. To better understand this process, we analyze JEV-guided retrieval from two complementary perspectives: (1) the extent to which JEV alters the retrieved demonstration set, and (2) whether these changes improve the utility of the selected demonstrations.

\paragraph{Demonstration replacement ratio.} Fig.~\ref{fig:jev_ratio} illustrates the distribution of the number of demonstrations replaced by different JEV models across varying shot settings. Both JEV-9B and NeoJEV consistently replace a substantial proportion of the demonstrations initially retrieved by $k$NN. As $K$ increases, the number of replaced demonstrations generally increases, indicating that JEV-guided reranking substantially modifies the original retrieval results. These findings reveal a notable discrepancy between semantic retrieval and JEV's assessment of demonstration usefulness, suggesting that semantically similar examples are not necessarily the most informative for ICL.

\paragraph{Replacement utility gain.} To assess whether JEV selects more useful demonstrations than those discarded from $k$NN, we introduce Replacement Utility Gain (RUG). Let $u(x_i,x_q)\in\{0,1\}$ indicate whether the LLM correctly predicts query $x_q$ using demonstration $x_i$ alone. Given the previously defined $\mathcal{C}_q^{s}$ and $\mathcal{C}_q^{Jev}$, we define RUG as 
\begin{equation}
\resizebox{0.98\columnwidth}{!}{$\displaystyle
\mathrm{RUG}(x_q)
=
\frac{\sum_{x_i\in\mathcal{C}_q^{Jev}\setminus\mathcal{C}_q^{s}}u(x_i,x_q)}
{|\mathcal{C}_q^{Jev}\setminus\mathcal{C}_q^{s}|}
-
\frac{\sum_{x_i\in\mathcal{C}_q^{s}\setminus\mathcal{C}_q^{Jev}}u(x_i,x_q)}
{|\mathcal{C}_q^{s}\setminus\mathcal{C}_q^{Jev}|}
$}.
\end{equation}
Here $|\mathcal{C}_q^{s}\setminus\mathcal{C}_q^{Jev}|$ denotes the number of demonstrations discarded from the original $k$NN retrieval results, which equals the number of demonstrations introduced by JEV.

Fig.~\ref{fig:jev_qua} shows that JEV-selected demonstrations generally become more useful relative to discarded $k$NN demonstrations as $K$ increases. The gains are more consistent on Qwen3-1.7B, whereas Llama3.2-1B exhibits task-dependent improvements, with positive effects on TREC, SST-5, and AG News but negative effects on SST-2 and MNLI. Notably, the slight decline at $K=8$ may be attributed to the fixed candidate pool size ($M=20$), which limits the availability of additional informative demonstrations as $K$ increases. Overall, these results suggest that JEV-guided replacement generally improves demonstration utility across most datasets, particularly under larger demonstration budgets, although its effectiveness remains dependent on the LLMs and task characteristics.

Taken together, these experiments demonstrate that Jev-guided reranking can identify more informative demonstrations than conventional semantic similarity-based retrieval, thereby improving overall ICL performance. Although its effectiveness varies across downstream models and tasks, these findings highlight the potential of Jev as a utility-aware demonstration evaluator.

\section{Method: Multi-objective Distillation}
\label{sec:method}

\begin{figure*}
    \centering
    \includegraphics[width=1\linewidth]{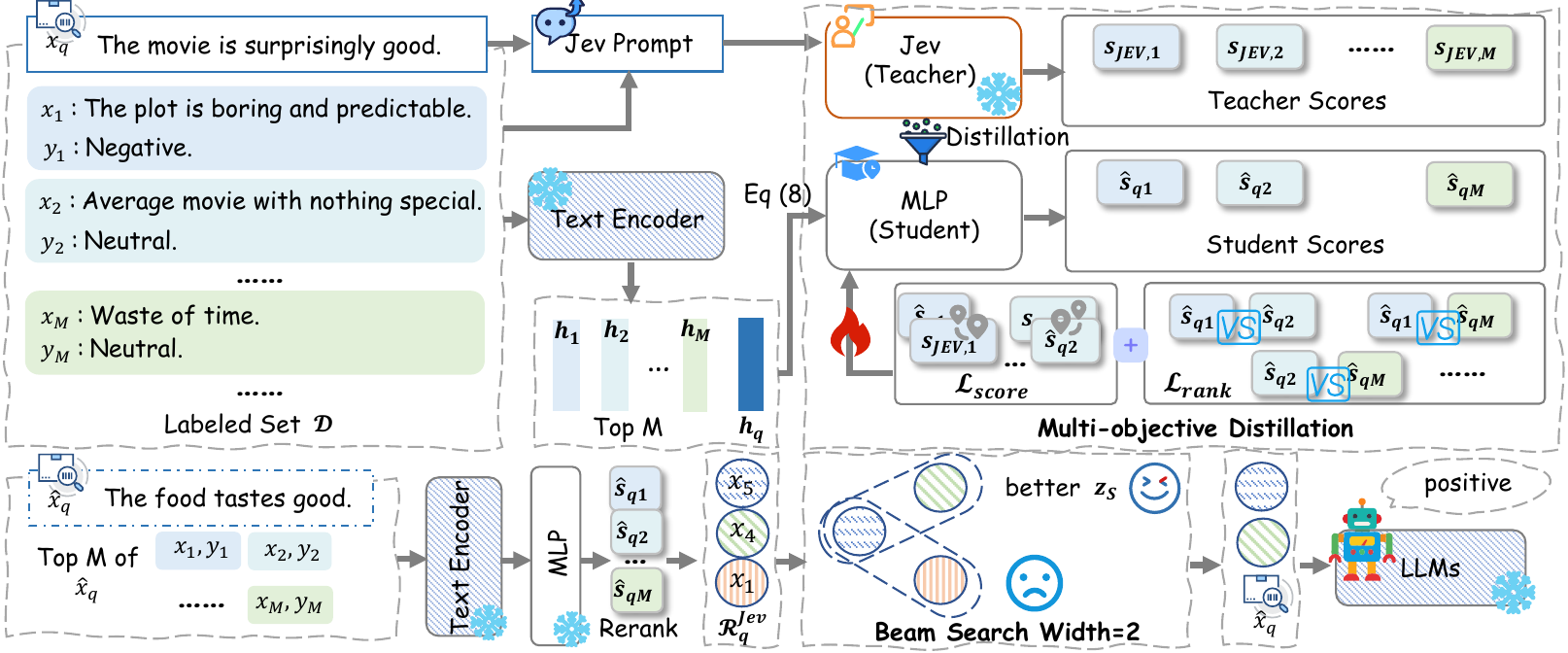}
    \caption{Overview of the proposed framework.}
    \label{fig:model}
\end{figure*}

Although JEV-guided retrieval improves the quality of in-context demonstrations, its inference overhead remains substantial. Given a candidate pool $\mathcal{D}_q$ containing $M$ demonstrations, JEV reranking requires $M$ forward evaluations. Motivated by knowledge distillation~\cite{xu2024survey}, we propose a multi-objective distillation framework as shown in Fig. \ref{fig:model} that transfers the utility estimation and ranking capabilities of JEV into a lightweight MLP student, enabling efficient demonstration selection without invoking JEV during inference.

\subsection{Distillation Data Construction}

To faithfully distill the teacher's behavior, we construct a JEV-labeled candidate set for each query $x_q$ from validation set. In practice, the validation set is split from the training set and is kept strictly disjoint from the test set. Specifically, given $M$ retrieved candidates in $\mathcal{R}_q$, we evaluate each candidate $x_i$ by Eq.~\ref{eq:jev_score}. These scores provide supervision for distilling the teacher's scoring and ranking behavior into a lightweight student model.

\subsection{Lightweight Utility Estimation}

We employ a frozen semantic encoder $g(\cdot)$ to obtain the representations of the query, candidate demonstrations, and their labels. Let $\mathbf{q}=g(x_q)$, $\mathbf{h}_i=g(x_i)$, and $\mathbf{l}_i=g(y_i)$, where $y_i$ denotes the natural language description of the candidate label. We construct a query-aware feature vector:
\begin{equation}
\mathbf{z}_i=[\mathbf{q};\mathbf{h}_i;|\mathbf{q}-\mathbf{h}_i|;\mathbf{q}\odot\mathbf{h}_i;sim_i;\mathbf{l}_i],
\end{equation}
where $sim_i$ denotes the cosine similarity between the query and candidate, $\odot$ represents element-wise multiplication, and $[\cdot;\cdot]$ denotes concatenation. Then, a lightweight MLP estimates the individual utility score as
\begin{equation} \label{eq:hat_s}
\hat{s}(x_q, x_i, y_i)=\sigma(MLP(\mathbf{z}_i)),
\end{equation}
where $\sigma(\cdot)$ is the sigmoid function.

\subsection{Multi-objective Distillation}

We optimize the student through two complementary objectives: score distillation and ranking distillation. The former preserves the teacher's utility estimates, while the latter transfers its relative preferences among candidate demonstrations.

\paragraph{Score Distillation.}
Given teacher scores $s_{\mathrm{Jev}}(x_q,x_i,y_i)$ and student predictions $\hat{s}(x_q,x_i,y_i)$, we minimize a soft-label binary cross-entropy:
\begin{equation}
\resizebox{0.98\columnwidth}{!}{$\displaystyle
\mathcal{L}_{\mathrm{score}} =
-\frac{1}{M}
\sum_{i=1}^{M}
\left[
s_{\mathrm{Jev}}\log\hat{s}
+(1-s_{\mathrm{Jev}})\log(1-\hat{s})
\right]
$}
\end{equation}
where $s_{\mathrm{Jev}}$ and $\hat{s}$ represent the teacher's utility score and the student's predicted score for each candidate, respectively.

\paragraph{Ranking Distillation.}
Predicting individual demonstration scores may overlook the relative differences between candidates. We therefore introduce a pairwise ranking objective based on RankNet~\cite{burges2005learning} to explicitly capture the teacher's preferences. For any two candidate demonstrations $x_i$ and $x_j$, we define $\Delta_{ij}=s_{\mathrm{Jev},i}-s_{\mathrm{Jev},j}$ and $\hat{\Delta}_{ij}=MLP(\mathbf{z}_i)-MLP(\mathbf{z}_j)$ as the teacher score difference and student logit difference, respectively. The pairwise ranking loss is defined as:
\begin{equation}
\ell_{iq}=\log\left(1+\exp\left[-\operatorname{sign}(\Delta_{ij})\hat{\Delta}_{ij}\right]\right).
\end{equation}
Here, $\operatorname{sign}(\Delta_{ij})$ indicates the teacher's preference direction, taking $+1$ if candidate $x_i$ is preferred over $x_j$ and $-1$ otherwise. To emphasize challenging comparisons, we introduce a difficulty-aware reweighting strategy that dynamically assigns greater importance to candidate pairs with larger ranking errors. The ranking objective is:
\begin{equation}
\mathcal{L}_{\mathrm{rank}}=
\frac{\sum_{(i,j)\in\mathcal{V}_q}\omega_{ij}\ell_{ij}}
{\sum_{(i,j)\in\mathcal{V}_q}\omega_{ij}},
\end{equation}
where $\mathcal{V}_q$ denotes the set of valid candidate pairs constructed from $M$ retrieved demonstrations, excluding pairs with identical teacher scores, and $\omega_{ij}$ represents the difficulty-aware weight, defined as:
\begin{equation}
\omega_{ij}=
\frac{\exp(\ell_{ij})}
{\sum_{(u,v)\in\mathcal{V}_q}\exp(\ell_{uv})}.
\end{equation}
This strategy emphasizes challenging pairs while retaining supervision from all valid pairs.

\paragraph{Joint Optimization.}
To jointly preserve the teacher's utility estimation and ranking capabilities, we combine the score distillation and pairwise ranking objectives $\mathcal{L}=\mathcal{L}_{\mathrm{score}} + \mathcal{L}_{\mathrm{rank}}.$

\subsection{Efficient Demonstration Selection}
During inference, given a test query $x_q$\footnote{With a slight abuse of notation, we use $x_q$ to denote the query instance, while emphasizing that the queries used during inference differ from those used during training.}, we first retrieve $M$ candidate demonstrations using semantic similarity and compute their representations with the frozen encoder. The distilled student then estimates individual utilities in a single batched forward pass by Eq.(\ref{eq:hat_s}), allowing candidates to be ranked without invoking JEV. Here, the candidate set reranked by the student is denoted as $\mathcal{R}_q^{Jev}$.

To further account for the interactions among demonstrations, we introduce a beam search strategy with a beam width of 4~\cite{song2025counterfactual}. Starting from an empty context, we iteratively expand each beam by adding candidate demonstrations from $\mathcal{R}_q^{Jev}$ and evaluate the resulting combinations using the distilled student. To accommodate multiple demonstrations within the MLP input, we construct a position-weighted representation for each ordered combination $Com=(x_1,\ldots,x_k)$, assigning greater importance to demonstrations closer to the query:
\begin{equation}
\resizebox{0.98\columnwidth}{!}{$\displaystyle
z_S=
[\mathbf{q};\bar{\mathbf{h}}_S;
|\mathbf{q}-\bar{\mathbf{h}}_S|;
\mathbf{q}\odot\bar{\mathbf{h}}_S;
\cos(\mathbf{q},\bar{\mathbf{h}}_S);
\bar{\mathbf{l}}_S],
$}
\end{equation}
where $\bar{\mathbf{h}}_S=\operatorname{Normalize}(\sum_{t=1}^{k}w_t\mathbf{h}_t)$ and $\bar{\mathbf{l}}_S=\operatorname{Normalize}(\sum_{t=1}^{k}w_t\mathbf{l}_t)$ are the aggregated demonstration and label representations, respectively, with $w_t=t/\sum_{j=1}^{k}j$ denoting the position-dependent weight. The resulting representation is evaluated by the same distilled student.

At each step, only the four combinations with the highest score are retained for subsequent expansion. Finally, the combination with the highest predicted utility is selected as the in-context demonstrations, rather than simply concatenating individually top-ranked candidates. The selected demonstrations are then incorporated into the prompt for in-context learning. Since demonstration selection is independent of the downstream LLM, our framework can be seamlessly applied to different LLMs without requiring model-specific training or modifications.

\section{Experiment}
\label{sec:exp}

\subsection{Experimental Setup}
\textbf{Details.} We conduct experiments on five text classification datasets: TREC, SST-2, SST-5, AG News, and MNLI. For each dataset, we sample 15\% of the original training set (up to 2,000 examples) as queries for distillation. Each query is paired with 32 candidates retrieved using all-MiniLM-L6-v2 and scored by JEV-9B. A shared MLP student with two hidden layers (768 and 256 dimensions) is trained across all datasets using score and ranking distillation. We optimize the student for 50 epochs using AdamW with a learning rate of $3\times10^{-4}$ and batch size 64. During inference top 20 samples from train set are used for 4-shot demonstration selection via beam search (beam width 4). We evaluate on Llama-3.2-1B, Qwen3-1.7B, Llama-3.1-8B, and Qwen3-8B, using up to 2,000 test examples per dataset. Classification accuracy is reported as the evaluation metric. All experiments are conducted on a single NVIDIA A100 80GB GPU. Further details are provided in App. \ref{app:details}.

\textbf{Baselines.} In addition to semantic $k$NN retrieval \cite{liu2022makes}, we compare our method with ConE \cite{peng2024revisiting} and DiSP \cite{wang2026easier}. \textbf{ConE} adopts a select-then-rerank strategy, retrieving semantically similar demonstrations and reranking candidate contexts based on the target LLM's conditional entropy. Lower-entropy contexts are preferred. For fair comparison, we use a candidate pool of 20 and select four demonstrations. \textbf{DiSP} trains lightweight difficulty-aware routers and judges to predict ICL success, enabling budget-constrained demonstration selection. We train DiSP using up to 2,000 training queries per dataset and restrict its candidate pool to 20. All methods use four demonstrations and share the same evaluation settings.

\subsection{Main Results} 
\begin{table}[h]
\centering
\footnotesize
\setlength{\tabcolsep}{3pt}
\renewcommand{\arraystretch}{0.7}

\resizebox{\columnwidth}{!}{%
\begin{tabular}{@{}lrrrrrr@{}}
\toprule
\textbf{Method}
& \textbf{TREC}
& \textbf{SST-2}
& \textbf{SST-5}
& \textbf{AG News}
& \textbf{MNLI}
& \textbf{Avg.} \\
\midrule

\rowcolor{blue!9}
\multicolumn{7}{c}{\textbf{Llama3.2-1B}} \\

$k$NN
& 72.60 & 58.83 & 38.10 & 86.65 & 42.05 & 59.65 \\

ConE
& 75.20 & \textbf{60.55} & 34.15
& 85.90 & 40.65 & 59.29 \\

DiSP
& 69.40 & 55.77 & 39.10
& 87.64 & \textbf{45.42} & 59.47 \\

\rowcolor{black!5}
\textbf{DisNeoJEV}
& \textbf{80.60} & 57.91 & 41.85
& \textbf{91.05} & 45.00 & \textbf{63.28} \\

\rowcolor{black!5}
\textbf{DisJEV}
& 80.00 & 58.14 & \textbf{41.90}
& 88.88 & 44.29 & 62.64 \\

\midrule

\rowcolor{blue!9}
\multicolumn{7}{c}{\textbf{Qwen3-1.7B}} \\

$k$NN
& 72.40 & \textbf{90.25} & 44.30
& 90.30 & 69.95 & 73.44 \\

ConE
& 78.40 & 87.89 & 47.10
& 83.80 & 71.55 & 73.75 \\

DiSP
& 78.40 & 89.61 & \textbf{47.19}
& 85.82 & 72.36 & 74.68 \\

\rowcolor{black!5}
\textbf{DisNeoJEV}
& \textbf{88.20} & 90.14 & 44.80
& \textbf{91.70} & 73.55 & \textbf{77.68} \\

\rowcolor{black!5}
\textbf{DisJEV}
& 85.20 & 89.91 & 45.48
& 90.16 & \textbf{74.85} & 77.12 \\

\midrule

\rowcolor{blue!9}
\multicolumn{7}{c}{\textbf{Llama3.1-8B}} \\

$k$NN
& 89.40 & 93.92 & \textbf{57.65}
& 90.64 & 60.74 & 78.47 \\

ConE
& 86.00 & 94.61 & 53.75
& 90.60 & 54.80 & 75.95 \\

DiSP
& 83.20 & 92.01 & 50.43
& 89.12 & \textbf{65.52} & 76.06 \\

\rowcolor{black!5}
\textbf{DisNeoJEV}
& \textbf{93.20} & 94.61 & 56.80
& \textbf{91.60} & 61.70 & \textbf{79.58} \\

\rowcolor{black!5}
\textbf{DisJEV}
& 91.20 & \textbf{94.72} & 57.15
& 90.36 & 61.09 & 78.90 \\

\midrule

\rowcolor{blue!9}
\multicolumn{7}{c}{\textbf{Qwen3-8B}} \\

$k$NN
& 88.40 & \textbf{93.23} & 49.80
& \textbf{92.20} & 78.40 & 80.41 \\

ConE
& 91.00 & 88.94 & 50.40
& 88.80 & 79.25 & 79.68 \\

DiSP
& 88.70 & 91.67 & 50.62
& 90.78 & 79.06 & 80.17 \\

\rowcolor{black!5}
\textbf{DisNeoJEV}
& \textbf{93.80} & 91.51 & 52.00
& 92.15 & \textbf{82.50} & \textbf{82.39} \\

\rowcolor{black!5}
\textbf{DisJEV}
& 92.80 & 91.79 & \textbf{52.58}
& 91.16 & 82.12 & 82.09 \\

\rowcolor{blue!9}
\multicolumn{7}{c}{\textbf{DeepSeek-V4-Flash}} \\

$k$NN
& 94.40 & 95.00 & 58.00
& 90.60 & 82.00 & 84.00 \\

\rowcolor{black!5}
\textbf{DisNeoJEV}
& \textbf{96.20} & \textbf{96.00} & \textbf{60.00}
& 92.60 & \textbf{84.40} & \textbf{85.84} \\
\rowcolor{black!5}
\textbf{DisJEV}
& 94.40 & 95.40 & 58.00
& \textbf{93.40} & 83.60 & 84.96 \\

\midrule

\rowcolor{blue!9}
\multicolumn{7}{c}{\textbf{Qwen3.8-Flash}} \\

$k$NN
& 96.00 & 96.00 & 56.40
& 91.60 & 88.20 & 85.64 \\

\rowcolor{black!5}
\textbf{DisNeoJEV}
& \textbf{96.20} & 96.00 & 60.00
& \textbf{92.60} & 88.20 & \textbf{86.60} \\
\rowcolor{black!5}
\textbf{DisJEV}
& 95.60 & \textbf{96.40} & \textbf{60.40}
& 89.80 & \textbf{89.00} & 86.24 \\

\bottomrule
\end{tabular}%
}
\caption{Classification accuracy (\%) of different demonstration retrieval
methods at $K=4$. The best result for each model and dataset is highlighted
in bold.}
\label{tab:main_results}

\end{table}

As shown in Tab.~\ref{tab:main_results}, both DisJEV and DisNeoJEV consistently outperform conventional retrieval baselines across all four open-source LLMs. Compared with $k$NN, DisJEV improves average accuracy by 2.99, 3.68, 0.43, and 1.68 percentage points, respectively. In contrast, ConE and DiSP exhibit inconsistent improvements and even degrade performance on several LLMs. DisNeoJEV further improves upon DisJEV, achieving average accuracies of 63.28\%, 77.68\%, 79.58\%, and 82.39\%, highlighting the importance of teacher discrimination quality. These results suggest that distilling demonstration utility knowledge provides a more effective and robust retrieval strategy than conventional alternatives. Moreover, the lightweight and label-agnostic student enables efficient beam search over demonstration combinations and joint training across diverse tasks. We further investigate computational efficiency in \S~\ref{sec:time} and cross-task generalization in \S~\ref{sec:abl}.

\paragraph{Generalization to Larger LLMs.}
We further evaluate both methods on DeepSeek-V4-Flash and Qwen3.8-Flash via APIs on 500 samples, excluding ConE and DiSP due to the difficulty and cost of repeated confidence estimation. DisJEV improves average accuracy over $k$NN by 0.96 and 0.60 percentage points, respectively, while DisNeoJEV achieves larger gains of 1.84 and 0.96 points. Although the improvements are relatively modest, these results confirm that utility-aware retrieval remains beneficial for stronger LLMs and that better teacher discrimination can further enhance student performance.

\subsection{Time Cost}  \label{sec:time}

\begin{figure}
    \centering
    \includegraphics[width=1\linewidth]{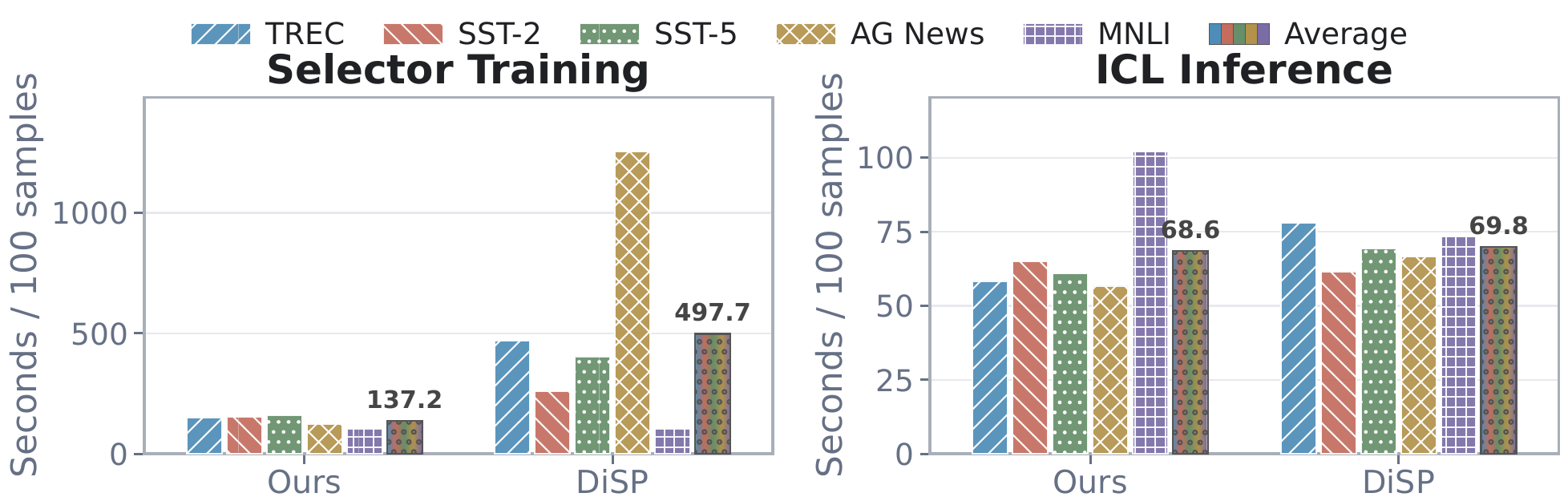}
    \caption{Time cost comparison between DisJEV and DiSP during training and inference.}
    \label{fig:time}
\end{figure}
We further compare the computational efficiency of DisJEV and DiSP using Qwen3-1.7B. Since JEV-9B has a larger number of parameters, DisJEV is expected to incur a higher computational cost. As shown in Fig.~\ref{fig:time}, DisJEV incurs substantially lower training costs than DiSP (137.2 vs. 497.7 seconds per 100 samples on average), primarily because our method employs a lightweight MLP student, whereas DiSP relies on a BERT-based selector. During ICL inference, DisJEV also achieves slightly lower average latency (68.6 vs. 69.8 seconds per 100 samples), outperforming DiSP in efficiency on four of the five datasets, with MNLI being the only exception. These results demonstrate that DisJEV achieves superior demonstration selection performance with significantly lower training costs and comparable or slightly better inference efficiency, offering a favorable trade-off between effectiveness and computational overhead.

\subsection{Ablation Study}  \label{sec:abl}

\begin{table}[!t]
\centering
\footnotesize
\setlength{\tabcolsep}{3.2pt}
\renewcommand{\arraystretch}{0.7}

\resizebox{\columnwidth}{!}{%
\begin{tabular}{@{}lrrrrrr@{}}
\toprule
\textbf{Method}
& \textbf{TREC}
& \textbf{SST-2}
& \textbf{SST-5}
& \textbf{AG News}
& \textbf{MNLI}
& \textbf{Avg.} \\
\midrule

\rowcolor{blue!9}
\multicolumn{7}{c}{\textbf{Llama3.2-1B}} \\

$k$NN
& 72.60 & 58.83 & 38.10 & 86.65 & 42.05 & 59.65 \\

DisJEV-trg
& 76.60 & 57.80 & \textbf{42.40}
& 86.55 & 42.70 & 61.21 \\

DisJEV-score
& 76.05 & \textbf{58.94} & 40.86
& 88.80 & 42.69 & 61.47 \\

DisJEV-rank
& 75.17 & 58.19 & 42.13
& 87.43 & 43.44 & 61.27 \\

\rowcolor{black!5}
\textbf{DisJEV}
& \textbf{80.00} & 58.14 & 41.90
& \textbf{88.88} & \textbf{44.29} & \textbf{62.64} \\

\midrule

\rowcolor{blue!9}
\multicolumn{7}{c}{\textbf{Qwen3-1.7B}} \\

$k$NN
& 72.40 & 90.25 & 44.30 & 90.30 & 69.95 & 73.44 \\
DisJEV-trg
& 78.80 & \textbf{90.48} & 44.95
& \textbf{90.60} & 72.25 & 75.42 \\
DisJEV-score
& 81.40 & 89.68 & 44.70
& 89.85 & 72.90 & 75.71 \\
DisJEV-rank
& 79.60 & 89.45 & 43.85
& 89.60 & 74.11 & 75.32 \\
\rowcolor{black!5}
\textbf{DisJEV}
& \textbf{85.20} & 89.91 & \textbf{45.48}
& 90.16 & \textbf{74.85} & \textbf{77.12} \\
\bottomrule
\end{tabular}%
}

\vspace{2pt}
\caption{Ablation study of DisJEV at $K=4$. The best results are highlighted in bold.}
\label{tab:ablation}
\end{table}

To investigate the effectiveness and generalization capability of DisJEV, we consider three variants: (1) \textbf{DisJEV-score}, which removes the ranking distillation loss and retains only score distillation; (2) \textbf{DisJEV-rank}, which removes the score distillation loss and retains only ranking distillation; and (3) \textbf{DisJEV-trg}, which adopts a leave-one-dataset-out strategy, training the student on four datasets and evaluating it on the held-out dataset without any target-task distillation.

As shown in Tab.~\ref{tab:ablation}, removing either distillation objective leads to a decline in average accuracy, indicating that score estimation and preference ranking provide complementary supervision. The complete DisJEV achieves the best average performance on both Llama3.2-1B and Qwen3-1.7B. More importantly, DisJEV-trg achieves average accuracies of 61.21\% and 75.42\%, respectively, outperforming $k$NN by 1.56 and 1.98 percentage points despite having no access to the target dataset during distillation. Compared with the fully trained DisJEV, the performance decreases by only 1.43 and 1.70 percentage points, respectively. These results demonstrate that the distilled utility estimator can transfer its learned knowledge to unseen tasks, supporting the cross-task generalization capability.

\subsection{Ranking Consistency}

\begin{figure}
    \centering
    \includegraphics[width=1\linewidth]{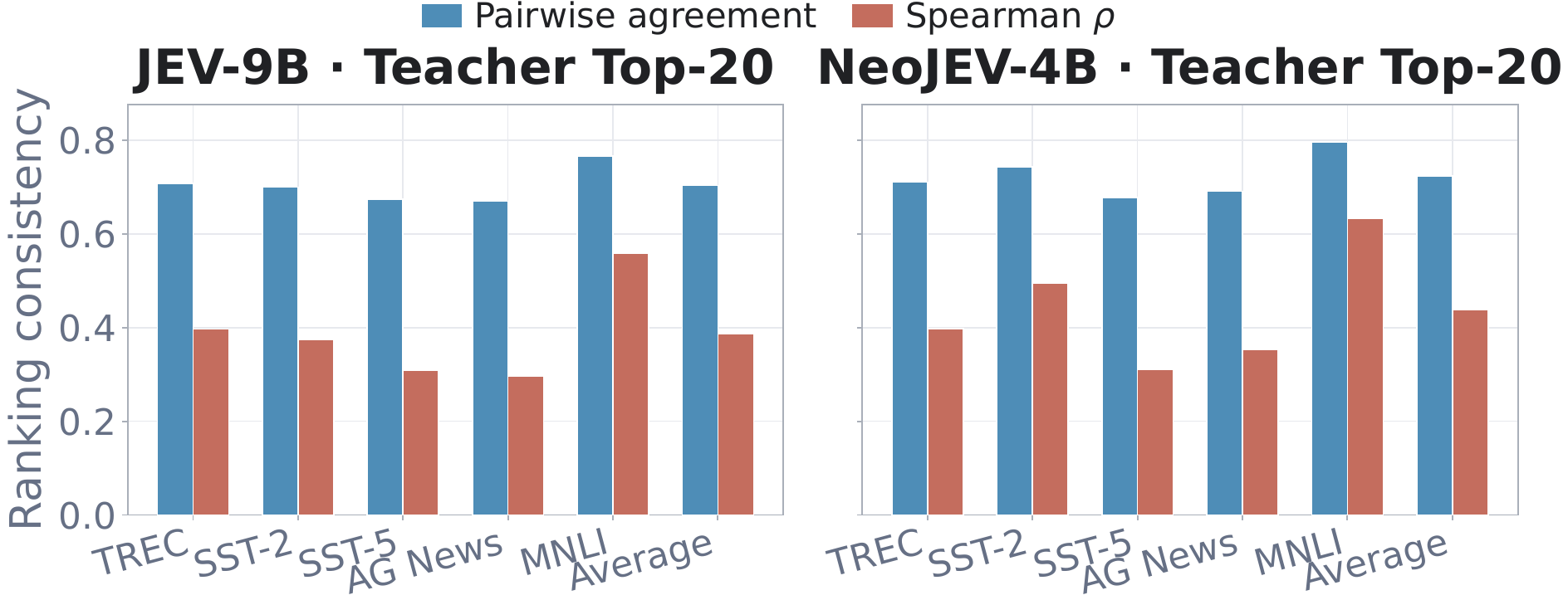}
    \caption{}
    \label{fig:ranking}
\end{figure}
We further evaluate the ranking consistency between the student and teacher models on the test sets using pairwise agreement and Spearman's rank correlation under $M=20$. As shown in Fig.~\ref{fig:ranking}, both DisJEV and DisNeoJEV achieve approximately 70\% pairwise agreement with their respective teachers, indicating that the distilled students effectively preserve the teachers' relative preferences among candidate demonstrations. Notably, DisNeoJEV generally achieves higher ranking consistency than DisJEV, particularly on SST-2 and MNLI, suggesting more effective preference transfer from NeoJEV-4B. Although Spearman's correlations are relatively moderate, the consistently positive values across all datasets further support the effectiveness of our multi-objective distillation in transferring demonstration utility knowledge.

\subsection{Beam Width}
\begin{figure}
    \centering
    \includegraphics[width=1\linewidth]{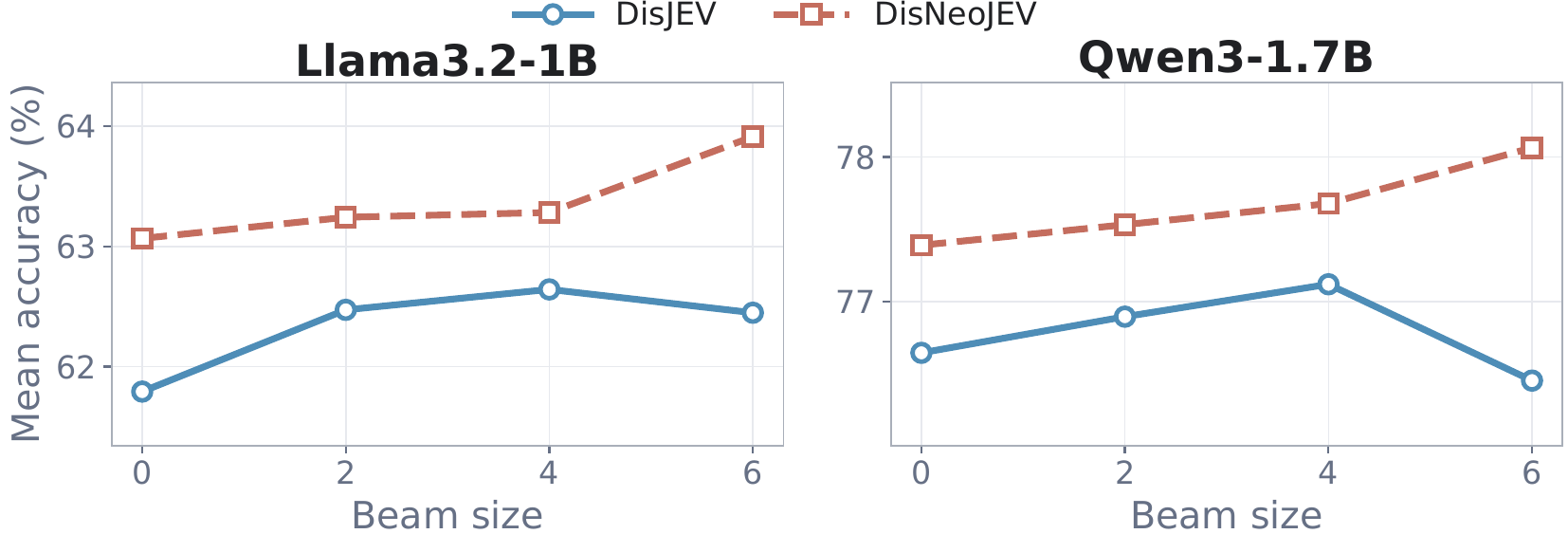}
    \caption{Effect of beam size on the average ICL accuracy of DisJEV and DisNeoJEV.}
    \label{fig:beam}
\end{figure}

To investigate the impact of beam search width, we evaluate DisJEV and DisNeoJEV under different beam sizes. As shown in Fig.~\ref{fig:beam}, the two methods exhibit distinct sensitivities to beam width. DisNeoJEV consistently benefits from larger beam sizes, with performance continuing to improve at a beam width of 6. In contrast, DisJEV achieves its best performance at a beam width of 4, beyond which further expansion leads to performance degradation. Considering the trade-off between retrieval effectiveness and computational cost, we uniformly set the beam width to 4 in all experiments.

\section{Conclusion}

In this paper, we investigate the potential of Jev for improving demonstration retrieval in ICL. Our findings show that Jev can identify informative demonstrations beyond semantic similarity, but its computational overhead limits practical deployment. To address this, we propose DisJEV, a lightweight and task-agnostic framework that distills Jev's utility estimation and ranking capabilities into a compact MLP. Extensive experiments demonstrate that DisJEV improves retrieval effectiveness and efficiency across diverse tasks and LLMs, while exhibiting promising cross-task generalization. Our study highlights the potential of decision-oriented model distillation for efficient and generalizable demonstration retrieval.

\clearpage
\bibliography{custom}

\appendix

\clearpage

\section{Details of Datasets and Settings} 
\label{app:details}

\textbf{Datasets}. Following the experimental benchmarks adopted by Wang et al.~\cite{wang2026easier}, we evaluate our framework on five widely used text classification datasets, covering question classification, sentiment analysis, topic classification, and natural language inference.

\paragraph{TREC.}
TREC is a question classification benchmark that requires models to identify the semantic type of a given question. We adopt the six coarse-grained categories: abbreviation, entity, description, human, location, and numeric.

\paragraph{SST-2 and SST-5.}
The Stanford Sentiment Treebank provides two sentiment classification benchmarks. SST-2 formulates sentiment analysis as binary classification with positive and negative labels, whereas SST-5 introduces five fine-grained sentiment categories ranging from very negative to very positive. The latter requires distinguishing subtle differences in sentiment intensity.

\paragraph{AGNews.}
AGNews is a news topic classification dataset consisting of news articles categorized into four topics: world, sports, business, and science/technology. The task requires models to identify the primary topic of each input text.

\paragraph{MNLI.}
MNLI is a natural language inference benchmark containing premise--hypothesis pairs from diverse genres. Given a premise and a hypothesis, the model predicts one of three semantic relationships: entailment, neutral, or contradiction.

\paragraph{Evaluation Protocol.}
For each dataset, we construct a labeled demonstration pool from the training split and evaluate in-context learning on held-out examples. Demonstration selection is performed without accessing the ground-truth labels of evaluation queries. We use dataset-specific label verbalizers and prompt templates, and report classification accuracy as the primary evaluation metric. The specific prompt templates used for each dataset are provided below.

\tcbset{
    compacticl/.style={
        enhanced,
        breakable,
        colback=gray!3,
        colframe=gray!45,
        colbacktitle=gray!15,
        coltitle=black,
        fonttitle=\bfseries\footnotesize,
        fontupper=\footnotesize,
        boxrule=0.4pt,
        arc=1pt,
        boxsep=1pt,
        left=5pt,
        right=5pt,
        top=3pt,
        bottom=3pt,
        before skip=4pt,
        after skip=4pt,
        boxed title style={boxrule=0pt,arc=1pt}
    }
}

\begin{tcolorbox}[compacticl,title={TREC ICL Input}]
\textbf{Task:} Classify the question type.\\
\textbf{Labels:} \texttt{abbreviation, description, entity, human,
location, numeric}.\\
\textbf{Instruction:} Use the examples and output exactly one label.\\[2pt]
\textbf{Example \(i\), \(i=1,\ldots,K\):}\\
\textbf{Question:} \texttt{[Demonstration question]}\\
\textbf{Label:} \texttt{[Demonstration label]}\\[2pt]
\textbf{Query:} \textbf{Question:} \texttt{[Target question]}\\
\textbf{Label:}
\end{tcolorbox}

\begin{tcolorbox}[compacticl,title={SST-2 ICL Input}]
\textbf{Task:} Classify the sentiment of the sentence.\\
\textbf{Labels:} \texttt{negative, positive}.\\
\textbf{Instruction:} Use the examples and output exactly one label.\\[2pt]
\textbf{Example \(i\), \(i=1,\ldots,K\):}\\
\textbf{Sentence:} \texttt{[Demonstration sentence]}\\
\textbf{Label:} \texttt{[Demonstration label]}\\[2pt]
\textbf{Query:} \textbf{Sentence:} \texttt{[Target sentence]}\\
\textbf{Label:}
\end{tcolorbox}

\begin{tcolorbox}[compacticl,title={SST-5 ICL Input}]
\textbf{Task:} Classify the sentiment of the sentence.\\
\textbf{Labels:} \texttt{very negative, negative, neutral, positive,
very positive}.\\
\textbf{Instruction:} Use the examples and output exactly one label.\\[2pt]
\textbf{Example \(i\), \(i=1,\ldots,K\):}\\
\textbf{Sentence:} \texttt{[Demonstration sentence]}\\
\textbf{Label:} \texttt{[Demonstration label]}\\[2pt]
\textbf{Query:} \textbf{Sentence:} \texttt{[Target sentence]}\\
\textbf{Label:}
\end{tcolorbox}

\begin{tcolorbox}[compacticl,title={AG News ICL Input}]
\textbf{Task:} Classify the news topic.\\
\textbf{Labels:} \texttt{World, Sports, Business, Science and Technology}.\\
\textbf{Instruction:} Use the examples and output exactly one label.\\[2pt]
\textbf{Example \(i\), \(i=1,\ldots,K\):}\\
\textbf{Article:} \texttt{[Demonstration article]}\\
\textbf{Label:} \texttt{[Demonstration label]}\\[2pt]
\textbf{Query:} \textbf{Article:} \texttt{[Target article]}\\
\textbf{Label:}
\end{tcolorbox}

\begin{tcolorbox}[compacticl,title={MNLI ICL Input}]
\textbf{Task:} Determine the relation between the premise and hypothesis.\\
\textbf{Labels:} \texttt{entailment, neutral, contradiction}.\\
\textbf{Instruction:} Use the examples and output exactly one label.\\[2pt]
\textbf{Example \(i\), \(i=1,\ldots,K\):}\\
\textbf{Premise:} \texttt{[Demonstration premise]}\\
\textbf{Hypothesis:} \texttt{[Demonstration hypothesis]}\\
\textbf{Label:} \texttt{[Demonstration label]}\\[2pt]
\textbf{Query:} \textbf{Premise:} \texttt{[Target premise]}\\
\textbf{Hypothesis:} \texttt{[Target hypothesis]}\\
\textbf{Label:}
\end{tcolorbox}

\section{More Results of Preliminary Experiment}
\label{app:more_r1}

\begin{table*}[h]
\centering
\scriptsize
\setlength{\tabcolsep}{3.5pt}
\renewcommand{\arraystretch}{1}

\resizebox{\textwidth}{!}{%
\begin{tabular}{
@{}llcccccc
@{\hspace{14pt}}
llcccccc@{}}
\toprule
\multicolumn{8}{c}{\textbf{Llama Models}}
&
\multicolumn{8}{c}{\textbf{Qwen Models}}
\\
\cmidrule(r){1-8}
\cmidrule(l){9-16}
\textbf{Method} & \textbf{$K$}
& \textbf{TREC}
& \textbf{SST-2}
& \textbf{SST-5}
& \textbf{AG News}
& \textbf{MNLI}
& \textbf{Avg.}
&
\textbf{Method} & \textbf{$K$}
& \textbf{TREC}
& \textbf{SST-2}
& \textbf{SST-5}
& \textbf{AG News}
& \textbf{MNLI}
& \textbf{Avg.}
\\
\midrule

% =========================================================
% Llama3.2-1B / Qwen3-1.7B
% =========================================================
\rowcolor{blue!9}
\multicolumn{8}{c}{\textbf{Llama3.2-1B}}
&
\multicolumn{8}{c}{\textbf{Qwen3-1.7B}}
\\

Zero-shot & 0
& 40.00 & 89.68 & 34.05 & 74.15 & 41.80 & 55.94
&
Zero-shot & 0
& 53.60 & 81.54 & 31.00 & 33.95 & 50.45 & 50.11
\\
\midrule

$k$NN & 1
& 78.40 & \textbf{72.13} & \textbf{39.25}
& 74.00 & \textbf{39.35} & \textbf{60.63}
&
$k$NN & 1
& 65.00 & 87.96 & 40.40
& 83.95 & 67.05 & 68.87
\\

NeoJevICL & 1
& \textbf{80.60} & 65.94 & 38.50
& 74.70 & 37.55 & 59.46
&
NeoJevICL & 1
& 77.60 & 85.78 & \textbf{43.85}
& \textbf{84.80} & \textbf{70.30} & 72.47
\\

\rowcolor{black!6}
JevICL & 1
& 78.20 & 64.91 & 36.30
& \textbf{74.75} & 34.50 & 57.73
&
JevICL & 1
& \textbf{78.20} & \textbf{88.53} & 43.45
& 84.00 & 69.85 & \textbf{72.81}
\\

$k$NN & 2
& 64.20 & \textbf{56.31} & 33.80
& 82.95 & 34.40 & 54.33
&
$k$NN & 2
& 67.00 & \textbf{87.96} & 42.00
& 90.35 & 70.50 & 71.56
\\

NeoJevICL & 2
& \textbf{67.80} & 51.83 & \textbf{37.15}
& \textbf{84.45} & \textbf{36.65} & \textbf{55.58}
&
NeoJevICL & 2
& \textbf{86.00} & 87.04 & \textbf{46.20}
& \textbf{91.45} & 73.70 & \textbf{76.88}
\\

\rowcolor{black!6}
JevICL & 2
& 65.60 & 53.67 & 36.30
& 83.80 & 33.40 & 54.55
&
JevICL & 2
& 83.80 & \textbf{87.96} & 44.45
& 90.15 & \textbf{74.15} & 76.10
\\

$k$NN & 4
& 72.60 & \textbf{58.83} & 38.10
& 86.65 & 42.05 & 59.65
&
$k$NN & 4
& 72.40 & 90.25 & 44.30
& 90.30 & 69.95 & 73.44
\\

NeoJevICL & 4
& \textbf{85.00} & 53.56 & 43.60
& \textbf{90.90} & 43.90 & \textbf{63.39}
&
NeoJevICL & 4
& \textbf{92.00} & \textbf{91.63} & \textbf{50.75}
& \textbf{92.05} & 73.50 & \textbf{79.99}
\\

\rowcolor{black!6}
JevICL & 4
& 82.00 & 54.24 & \textbf{44.65}
& 88.70 & 41.15 & 62.15
&
JevICL & 4
& 89.60 & 91.17 & 49.05
& 90.80 & \textbf{74.05} & 78.93
\\

$k$NN & 6
& 77.60 & \textbf{61.01} & 38.90
& 87.95 & 46.30 & 62.35
&
$k$NN & 6
& 75.60 & 90.02 & 43.85
& 90.85 & 68.75 & 73.81
\\

NeoJevICL & 6
& 88.40 & 57.22 & \textbf{45.95}
& \textbf{92.05} & \textbf{48.55} & \textbf{66.43}
&
NeoJevICL & 6
& \textbf{90.20} & \textbf{92.78} & \textbf{50.85}
& \textbf{92.10} & 74.40 & \textbf{80.07}
\\

\rowcolor{black!6}
JevICL & 6
& \textbf{88.60} & 56.65 & 45.50
& 90.40 & 48.30 & 65.89
&
JevICL & 6
& 89.00 & 91.40 & 48.70
& 91.30 & \textbf{74.45} & 78.97
\\

$k$NN & 8
& 79.00 & \textbf{65.48} & 39.10
& 90.60 & 46.35 & 64.11
&
$k$NN & 8
& 76.80 & 88.76 & 44.45
& 91.25 & 67.95 & 73.84
\\

NeoJevICL & 8
& \textbf{89.20} & 61.12 & 44.80
& \textbf{92.00} & 49.95 & 67.41
&
NeoJevICL & 8
& \textbf{90.20} & \textbf{92.43} & \textbf{51.70}
& \textbf{92.15} & \textbf{74.15} & \textbf{80.13}
\\

\rowcolor{black!6}
JevICL & 8
& 87.00 & 61.58 & \textbf{45.65}
& 90.55 & \textbf{52.45} & \textbf{67.45}
&
JevICL & 8
& 87.40 & 91.51 & 48.85
& 91.35 & 72.35 & 78.29
\\

\midrule

% =========================================================
% Llama3.1-8B / Qwen3-8B
% =========================================================
\rowcolor{blue!9}
\multicolumn{8}{c}{\textbf{Llama3.1-8B}}
&
\multicolumn{8}{c}{\textbf{Qwen3-8B}}
\\

Zero-shot & 0
& 76.80 & 97.13 & 51.00 & 78.20 & 47.80 & 70.19
&
Zero-shot & 0
& 69.80 & 94.27 & 44.20 & 82.70 & 70.80 & 72.35
\\
\midrule

$k$NN & 1
& 81.20 & \textbf{96.33} & 55.50
& 85.85 & 52.75 & 74.33
&
$k$NN & 1
& 76.80 & 94.38 & 50.55
& 90.95 & 75.75 & 77.69
\\

NeoJevICL & 1
& 86.40 & 95.30 & \textbf{57.55}
& \textbf{86.30} & \textbf{61.20} & \textbf{77.35}
&
NeoJevICL & 1
& \textbf{90.40} & \textbf{94.50} & \textbf{52.90}
& \textbf{91.55} & 77.15 & \textbf{81.30}
\\

\rowcolor{black!6}
JevICL & 1
& \textbf{87.40} & 95.87 & 55.20
& 85.35 & 57.45 & 76.25
&
JevICL & 1
& 90.20 & 93.35 & 51.30
& 90.80 & \textbf{78.60} & 80.85
\\

$k$NN & 2
& 79.80 & \textbf{93.58} & 52.55
& 88.80 & 50.85 & 73.12
&
$k$NN & 2
& 83.00 & 93.23 & 50.85
& 91.75 & 78.15 & 79.40
\\

NeoJevICL & 2
& 92.40 & 93.23 & \textbf{56.35}
& \textbf{90.45} & \textbf{56.05} & \textbf{77.70}
&
NeoJevICL & 2
& \textbf{94.60} & \textbf{94.72} & \textbf{55.20}
& \textbf{92.00} & 80.50 & \textbf{83.40}
\\

\rowcolor{black!6}
JevICL & 2
& \textbf{92.60} & 93.46 & 55.80
& 88.75 & 55.40 & 77.20
&
JevICL & 2
& 92.80 & 92.89 & 52.55
& 90.95 & \textbf{81.40} & 82.12
\\

$k$NN & 4
& 78.20 & \textbf{94.72} & 54.25
& 89.80 & 57.05 & 74.80
&
$k$NN & 4
& 88.40 & \textbf{93.23} & 49.80
& 92.20 & 78.40 & 80.41
\\

NeoJevICL & 4
& \textbf{95.00} & 94.15 & \textbf{58.10}
& \textbf{91.70} & 60.25 & \textbf{79.84}
&
NeoJevICL & 4
& \textbf{95.80} & 93.81 & \textbf{54.80}
& \textbf{92.45} & \textbf{82.75} & \textbf{83.92}
\\

\rowcolor{black!6}
JevICL & 4
& 93.60 & 94.38 & 57.30
& 90.60 & \textbf{62.50} & 79.68
&
JevICL & 4
& 93.20 & 92.66 & 53.05
& 91.35 & 81.70 & 82.39
\\

$k$NN & 6
& 83.20 & \textbf{94.95} & 56.65
& 90.40 & 60.60 & 77.16
&
$k$NN & 6
& 91.00 & 93.69 & 49.70
& 91.90 & 79.90 & 81.24
\\

NeoJevICL & 6
& \textbf{94.40} & 94.84 & \textbf{60.35}
& \textbf{92.45} & 62.85 & \textbf{80.98}
&
NeoJevICL & 6
& \textbf{95.80} & \textbf{94.15} & \textbf{53.60}
& \textbf{92.45} & 81.45 & \textbf{83.49}
\\

\rowcolor{black!6}
JevICL & 6
& 94.20 & 94.61 & 60.20
& 91.25 & \textbf{64.40} & 80.93
&
JevICL & 6
& 93.60 & 93.69 & 52.85
& 91.60 & \textbf{82.90} & 82.93
\\

$k$NN & 8
& 84.80 & \textbf{95.18} & 56.40
& 90.35 & 63.30 & 77.99
&
$k$NN & 8
& 92.60 & 93.58 & 50.40
& 92.00 & 80.05 & 81.73
\\

NeoJevICL & 8
& \textbf{95.20} & 94.04 & \textbf{60.25}
& \textbf{92.45} & 65.30 & \textbf{81.45}
&
NeoJevICL & 8
& \textbf{94.20} & 93.69 & \textbf{53.55}
& \textbf{92.45} & 82.45 & \textbf{83.27}
\\

\rowcolor{black!6}
JevICL & 8
& 93.40 & 93.46 & 60.15
& 91.00 & \textbf{65.80} & 80.76
&
JevICL & 8
& 93.20 & \textbf{93.81} & 53.25
& 91.90 & \textbf{82.65} & 82.96
\\

\bottomrule
\end{tabular}%
}

\caption{Classification accuracy (\%) of $k$NN, NeoJevICL,
and JevICL across different numbers of demonstrations ($K$).
The best result for each model, shot setting, and dataset
is highlighted in bold.}
\label{tab:pre_results}
\end{table*}

Tab.~\ref{tab:pre_results} reports the classification accuracy of different demonstration retrieval methods across five datasets, four downstream LLMs, and varying numbers of demonstrations. Overall, JEV-guided reranking generally outperforms conventional $k$NN retrieval, suggesting that semantic similarity alone is insufficient to identify the most informative demonstrations. The improvements become more pronounced as the number of demonstrations increases. At $K=8$, JevICL improves the average accuracy over $k$NN by 3.34, 4.45, 2.77, and 1.23 percentage points on Llama3.2-1B, Qwen3-1.7B, Llama3.1-8B, and Qwen3-8B, respectively. These results demonstrate the potential of decision-oriented models to improve demonstration selection beyond semantic similarity.

Interestingly, NeoJevICL achieves comparable or even better performance than JevICL despite using a smaller 4B-parameter evaluator. At $K=8$, NeoJevICL achieves average accuracies of 67.41\%, 80.13\%, 81.45\%, and 83.27\% across the four downstream LLMs, outperforming JevICL on three of them. This observation is consistent with the stronger discriminative capabilities reported for NeoJEV-4B in its official evaluation, suggesting that the quality of decision-making supervision, rather than model size alone, is crucial for effective demonstration reranking.

We also observe that the effectiveness of JEV-guided retrieval varies across downstream models and demonstration budgets. For Llama3.2-1B, improvements are less stable in low-shot settings, whereas both JevICL and NeoJevICL generally provide more substantial gains when $K\geq4$. In contrast, the Qwen models exhibit more consistent improvements, particularly with NeoJevICL. Moreover, the relative gains tend to diminish for larger downstream LLMs, suggesting that stronger models may be less sensitive to demonstration quality. Nevertheless, the overall results highlight the effectiveness of decision-guided retrieval and motivate distilling such capabilities into a lightweight demonstration selector.

\subsection{Beam Size Details}
\begin{figure*}
    \centering
    \includegraphics[width=1\linewidth]{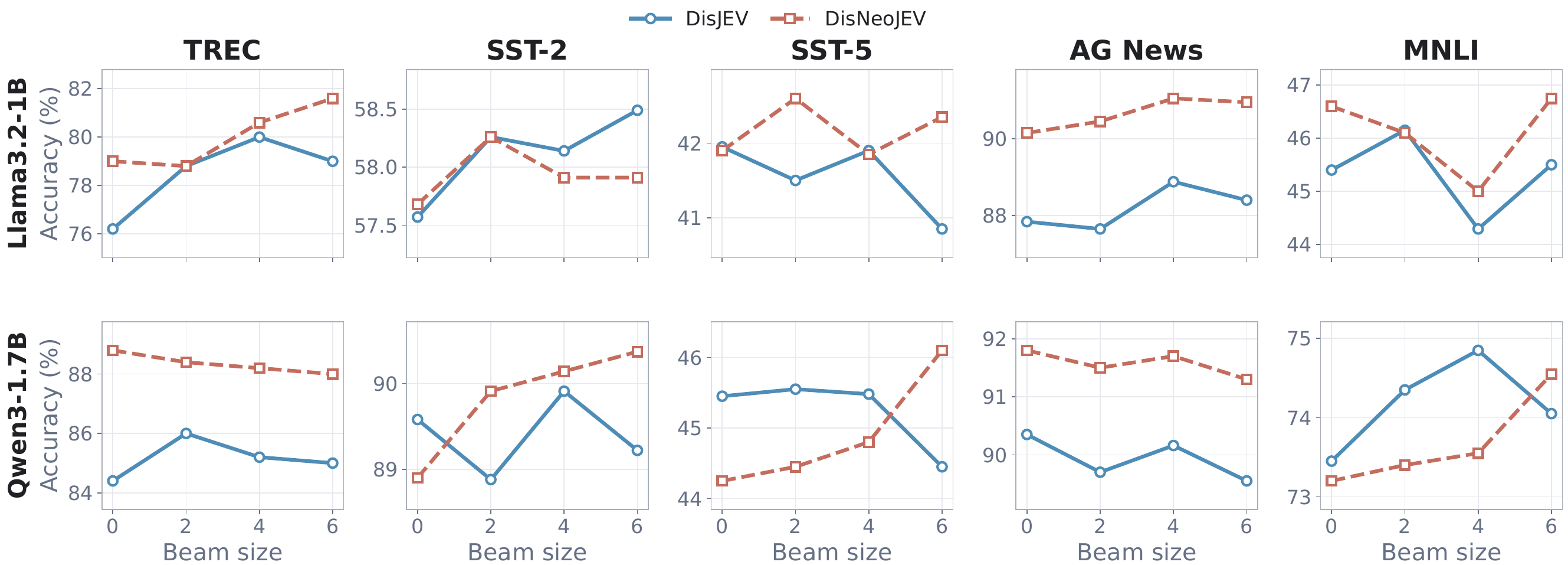}
    \caption{Detailed results across different beam sizes. }
    \label{fig:all_beam_size}
\end{figure*}
To investigate the detailed impact of beam width on demonstration selection, we evaluate DisJEV and DisNeoJEV across different datasets and LLMs. As shown in Fig.~\ref{fig:all_beam_size}, although the optimal beam width varies across datasets and teacher models, moderately increasing the beam width generally improves retrieval performance. In particular, DisNeoJEV often benefits from larger beam widths, whereas DisJEV tends to achieve its best performance at smaller widths. These results suggest that exploring more demonstration combinations is generally beneficial, although the gains are not strictly monotonic. Considering the overall performance and computational cost, we uniformly set the beam width to 4 in all experiments.

\end{document}